\documentclass[letterpaper, 10 pt, conference]{ieeeconf}  

\IEEEoverridecommandlockouts                              

\usepackage{amsmath}
\usepackage{morewrites}
\usepackage{etex}
\usepackage[utf8]{inputenc}
\usepackage{url}
\usepackage{booktabs}
\usepackage{amssymb}
\usepackage{bbding}
\usepackage{pifont}
\usepackage{wasysym}
\usepackage{utfsym}
\usepackage{fontawesome}
\usepackage{subcaption}
\usepackage{graphicx}
\usepackage{caption}
\usepackage{amssymb}
\usepackage{multirow} 
\usepackage{xcolor}

\title{\LARGE \bf
DNRSelect: Active Best View Selection for Deferred Neural Rendering
}

\author{Dongli Wu$^{1}$, Haochen Li$^{2}$, Xiaobao Wei$^{3\dagger}$
\thanks{$^{1}$Dongli Wu is with the 
 College of Design and Engineering, National University of Singapore. $^{2}$Haochen Li is with the School of Cyber Science and Technology, Beihang University. $^{3}$Xiaobao Wei is with the University of Chinese Academy of Sciences and Institute of Software, Chinese Academy of Sciences.}
\thanks{$\dagger$ Corresponding to weixiaobao0210@gmail.com.}
}

\begin{document}

\maketitle
\thispagestyle{empty}
\pagestyle{empty}

\begin{abstract}

Deferred neural rendering (DNR) is an emerging computer graphics pipeline designed for high-fidelity rendering and robotic perception. However, DNR heavily relies on datasets composed of numerous ray-traced images and demands substantial computational resources. It remains under-explored how to reduce the reliance on high-quality ray-traced images while maintaining the rendering fidelity. In this paper, we propose DNRSelect, which integrates a reinforcement learning-based view selector and a 3D texture aggregator for deferred neural rendering. We first propose a novel view selector for deferred neural rendering based on reinforcement learning, which is trained on easily obtained rasterized images to identify the optimal views. By acquiring only a few ray-traced images for these selected views, the selector enables DNR to achieve high-quality rendering. To further enhance spatial awareness and geometric consistency in DNR, we introduce a 3D texture aggregator that fuses pyramid features from depth maps and normal maps with UV maps. Given that acquiring ray-traced images is more time-consuming than generating rasterized images, DNRSelect minimizes the need for ray-traced data by using only a few selected views while still achieving high-fidelity rendering results. We conduct detailed experiments and ablation studies on the NeRF-Synthetic dataset to demonstrate the effectiveness of DNRSelect. The code will be released.

\end{abstract}

\section{INTRODUCTION}

Neural rendering and reconstruction offer essential solutions for 3D robotic perception by integrating traditional graphics techniques with neural networks. 
Recent approaches, such as Neural Radiance Fields (NeRF)~\cite{mildenhall2021nerf} and the more efficient 3D Gaussian Splatting (3DGS)~\cite{kerbl3Dgaussians}, have succeeded in novel view synthesis for scene reconstruction. To tackle the challenges of robotic exploration in novel environments, studies like ActiveNeRF~\cite{pan2022activenerf} and CG-SLAM~\cite{hu2024cg}, which are based on NeRF or 3DGS, have focused on the next best view (NBV) problem. These methods enable robots to actively determine the optimal perception strategy, facilitating comprehensive scene understanding.

Compared to NeRF and 3DGS for novel view synthesis, Deferred Neural Rendering (DNR)~\cite{DNR} integrates geometry-aware components, such as UV maps, into a differential rendering process to enhance object control and synthesis photorealism. Building on the DNR framework, several works~\cite{gao2020deferred, raj2021anr, worchel2022multi, wu2024deferredgs} have deferred the synthesis pipeline to improve shading effects and provide better scene editing capabilities. 
DNR enhances robotic visual perception by synthesizing high-quality images from incomplete or noisy sensor data, making it effective for tasks such as navigation, manipulation, and interaction in dynamic environments~\cite{tewari2022advances}. Additionally, by integrating geometry-aware components, DNR improves the accuracy and robustness of 3D scene understanding, enabling more precise and reliable performance in complex and unstructured settings. 
While DNR offers an effective approach to handling 3D reconstruction, it still faces two obvious limitations: 
1) \textbf{Heavy reliance on numerous ray-traced images.} DNR requires a large number of high-quality ray-traced images to produce detailed shading effects. However, acquiring such images is time-consuming and often impractical due to the significant computational resources needed for their generation. 
Meanwhile, an insufficient number of training images can cause the DNR renderer to overfit, resulting in noisy synthesis results. To date, no studies have specifically addressed reducing DNR's reliance on high-quality training datasets.
2) \textbf{Limited spatial awareness and geometric consistency.} Traditional neural rendering methods, including vanilla DNR, often struggle with maintaining geometric consistency and reducing artifacts, especially when dealing with sparse view inputs or insufficient geometric information.


\begin{figure}
    \centering
    \includegraphics[width=1\linewidth]{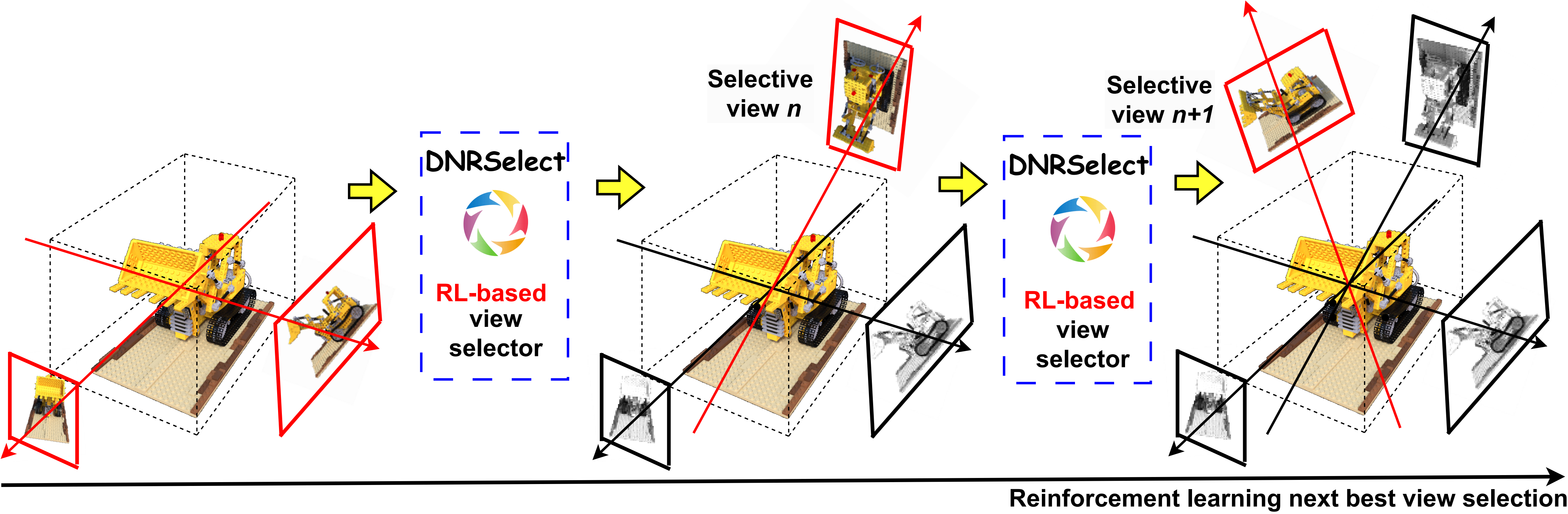}
    \caption{\textbf{DNRSelect}: We employ reinforcement learning to select a sequence of views that maximizes information gain, enhancing the quality of novel view synthesis with minimal additional resources, thereby aiding DNR in reconstruction.}
    \label{fig: intro}
    \vspace{-5mm}
\end{figure}
To address the above limitations, we propose DNRSelect, a novel deferred neural rendering framework that consists of a reinforcement learning-based (RL-based) view selector and a 3D texture aggregator. As shown in Fig.~\ref{fig: intro}, to reduce the reliance on high-quality ray-traced images, we introduce a view selector to choose the best views for better covering the target object. The view selector is trained along with DNR on easily obtained rasterized images in a reinforcement learning manner, which actively selects the optimal camera view combination according to the training state of DNR. To further mitigate the effects of sparse training views, we design a novel texture aggregator for 3D texture integration. Leveraging the proposed RL-based view selector and 3D texture aggregator, DNRSelect achieves high-fidelity rendering with a limited number of camera poses. Our main contributions can be summarized as follows:
\begin{itemize}
\item We propose DNRSelect, a novel deferred neural renderer that utilizes a reinforcement learning-based view selector to minimize the dependence on numerous ray-traced images by selecting optimal camera views for DNR optimization.

\item To address the sparse view problem and improve geometric consistency, we further present a learnable 3D texture aggregator that fuses pyramid features from depth maps, normal maps, and UV maps.

\item We perform comprehensive experiments on the NeRF-Synthetic dataset to illustrate the effectiveness of each component, revealing that DNRSelect consistently outperforms other neural rendering methods.

\end{itemize}

\section{Related Work}
\noindent\textbf{Neural Rendering.}
Neural rendering has recently gained significant attention in computer vision and graphics due to its diverse applications in capturing and rendering real-world scenes~\cite{DNR, tewari2020state, yan2024neural, wei2023medsam}. 
Early approaches, such as Neural Radiance Fields (NeRF)~\cite{mildenhall2021nerf} and its variants~\cite{muller2022instant, fridovich2022plenoxels, chen2022tensorf, wang2021neus, wei2024nto3d}, achieve realistic novel view synthesis and neural reconstruction by employing neural networks. However, these methods suffer from slow optimization and rendering speeds. To address this, 3D Gaussian Splatting~\cite{kerbl3Dgaussians} and its variants~\cite{cheng2024gaussianpro, huang2024S3Gaussian, lu2024scaffold}, which are point-based representations, have significantly accelerated the rendering process.
In contrast to these methods, which often lack explicit geometric control, deferred neural rendering (DNR)~\cite{DNR} and its variants~\cite{gao2020deferred, raj2021anr, wu2024deferredgs} introduce geometry-aware components, such as UV maps, into the rendering pipeline. This approach enables high-fidelity shading effects and provides greater flexibility in scene editing. 

In this paper, to address sparse view rendering in vanilla DNR, we propose DNRSelect and introduce a novel 3D texture aggregator that integrates multiple geometry-related features to enhance geometric consistency.



\noindent\textbf{Next Best View.}
The problem of Next Best View (NBV)~\cite{10610397, doi:10.1177/0278364911410755} has been extensively studied in robotic vision and computer graphics. 
NBV is a key aspect of active robot perception~\cite{doi:10.1177/0278364911410755, monica2019humanoid, jin2023neu, morrison2019multi, naazare2022online, hou2023learning}, enabling robots to actively adjust their sensors to capture the most useful informative data in a new environment. Several efforts have been made to tackle 3D reconstruction on robots. 
Based on uncertainty estimation for the entire scene, these works~\cite{pan2022activenerf, 10161012, goli2024bayes, xiao2024nerf} explore various NBV selection strategies using NeRF to reduce reliance on extensive training data and mitigate rendering artifacts caused by limited camera poses.

To the best of our knowledge, DNRSelect is the first approach to explore active deferred neural rendering using a reinforcement learning-based NBV strategy for optimal view selection. It enhances robotic 3D perception by effectively reducing the required data in complex environments.

\section{Method}

\subsection{Preliminary}

Deferred Neural Rendering (DNR)~\cite{DNR} enhances the graphics pipeline by replacing manual parameter tuning with neural networks for rendering, using neural textures that store high-dimensional features to enable realistic image synthesis. The neural texture $T$ generated by hierarchical sampling can preserve texture features in divergent resolution, which serves as Mipmaps. Then DNR utilizes a U-Net-like renderer $R$ to process screen-space feature maps, achieving view-dependent effects with spherical harmonics. Both the renderer $R$ and the neural texture $T$ are optimized end-to-end using a rendering loss $L$. For a training dataset with $N$ posed ray-traced images $\{I_{k},p_{p}\}^{N}_{k=1}$, the joint optimization of the renderer and the neural texture can be formulated as:
\vspace{-1.5mm}
\begin{equation}
   T^{*}, R^{*}=arg min\sum_{k=1}^{N}L(l_{k},p_{k}|T,R) 
\vspace{-1.5mm}
\end{equation}
where $L$ is a suitable training loss, such as a photometric loss. By training until convergence, we can obtain the optimal renderer $R^{*}$ and the most suitable neural texture $T^{*}$.


\subsection{Overall Framework}

\begin{figure*}[!ht]
    \centering
    \includegraphics[width=\linewidth]{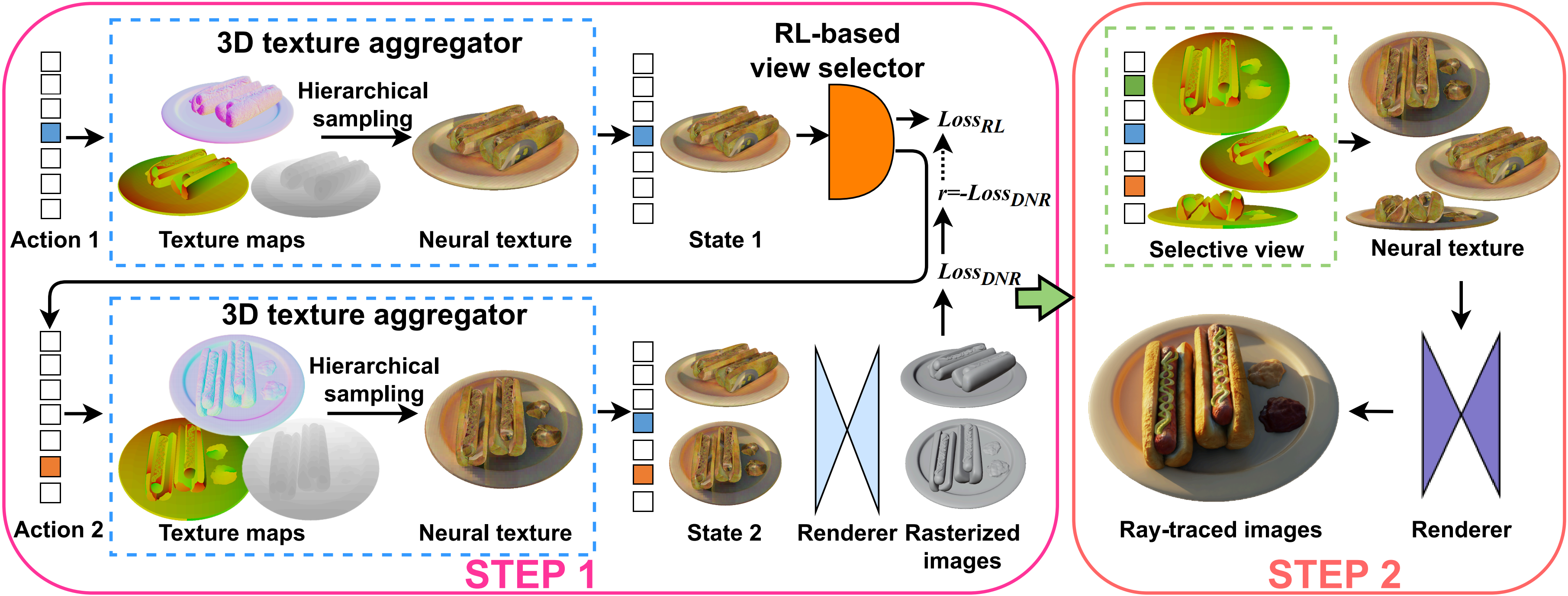}
    \caption{The overall pipeline of DNRSelect mainly includes two steps. In Step 1, we introduce a reinforcement learning-based (RL-based) view selector to minimize the reliance on abundant ray-traced images. Meanwhile, the DNR model learns the coarse geometric priors of the target object. In Step 2, the coarse DNR model is fine-tuned on the selected ray-traced images to achieve finer textures. Additionally, we enhance the hierarchical sampling in the vanilla DNR with a novel 3D texture aggregator, which accounts for the geometric information missing due to sparse selective views.}
    \label{fig:pipeline}
    \vspace{-6mm}
\end{figure*}

Fig.~\ref{fig:pipeline} outlines two primary steps in DNRSelect. Step 1 introduces a reinforcement learning-based view selector to identify the views containing the most informative content. The DNR loss function serves as a reward signal to guide the optimization of the view selector, which is trained using readily available rasterized images. Subsequently, DNR learns the coarse geometry of the target object, while the view selector produces a sequence of optimal views for the next stage. In Step 2, the parameters of DNR are fine-tuned using the selected viewpoints and their corresponding ray-traced images, thereby enabling DNR to achieve more refined texture synthesis.
To enhance geometric consistency despite the sparse views during fine-tuning, we employ a novel 3D texture aggregator to fuse depth maps, normal maps, and UV maps in two steps, which complements the 3D geometry prior.
The following subsections will provide a detailed introduction to the reinforcement learning-based view selector and the 3D texture aggregator.

\subsection{Reinforcement Learning-based View Selector}
\label{sec:rl-based}
\begin{figure}
    \centering
    \includegraphics[width=1.0\linewidth]{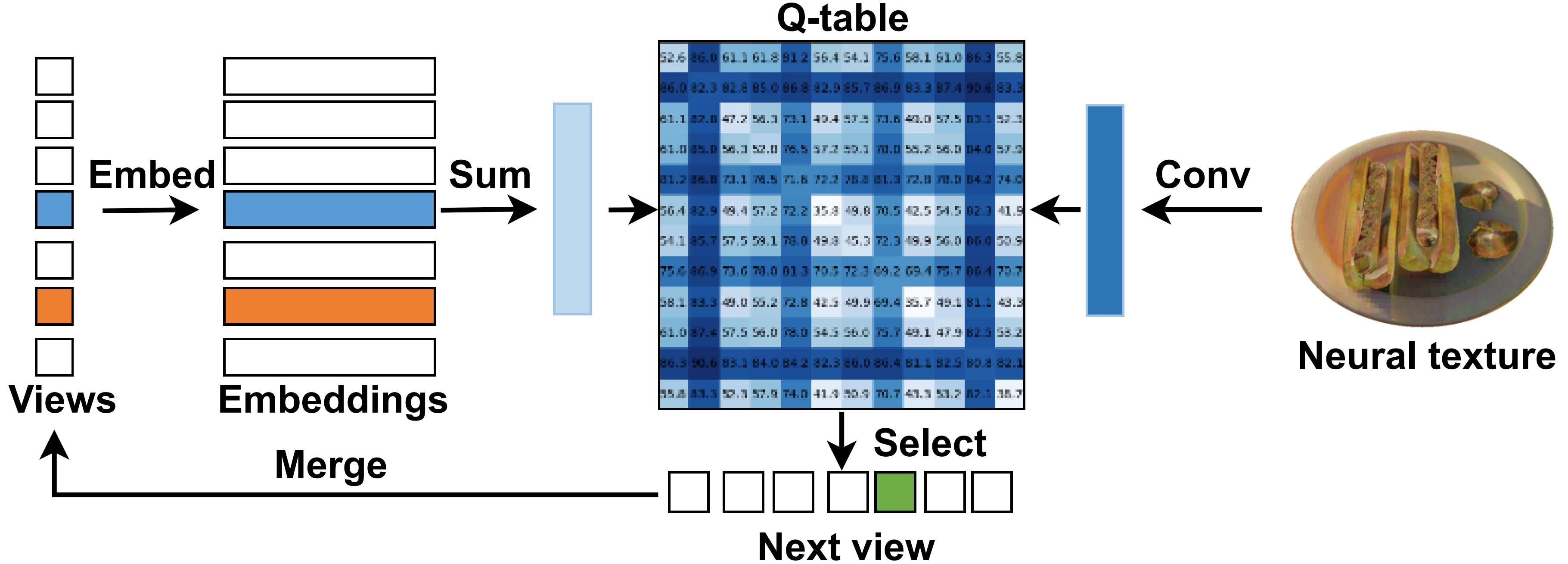}
    \caption{Illustration of the proposed RL-based view selector. 
    The selector encodes the selected views into embeddings and utilizes a convolutional neural network to transform the neural textures into features representing the current state. 
    Then, Q-table is employed to approximate the mapping between the current states and their corresponding values. 
    After optimization, the next best view is selected using a greedy algorithm that maximizes the expected reward.
    }
    \label{fig:view_selector}
    \vspace{-6mm}
\end{figure}


To alleviate the reliance on high-quality ray-traced images, we propose a reinforcement learning-based (RL-based) view selector in the multi-view DNR training inspired by ~\cite{hou2023learning}. 
The view selector dynamically selects the most suitable training views from \(N\) available options, ultimately choosing a total of \(M < N\) views. 
Details in the view selector are illustrated in Fig.~\ref{fig:view_selector}. 
Beginning with a random initial view \(a_{1}\), it iteratively selects subsequent views to optimize rendering quality while minimizing the number of views. 
This selection process is formulated as reinforcement learning, where the view selector acts as the agent and the DNR model serves as the environment. 
The selector's architecture consists of two branches: one transforms the view selection \(s_{t}^{cam}\) into a learnable camera embedding, while the other processes the neural texture \(s_{t}^{obs}\) into a hidden vector.
DNRSelect employs Q-learning~\cite{lillicrap2015continuous} to train the view selector, where the action value function \(Q(\cdot,\cdot)\) estimates the cumulative future reward for taking action \(a_{t+1}\) in the state \(s_{t}=\{s_{t}^{cam}, s_{t}^{obs}\}\):
\vspace{-2mm}
\begin{equation}
Q(s_{t},a_{t+1})=E(\sum_{\tau=t+1}^{T}\gamma^{\tau-t-1}r_{\tau})
\vspace{-2mm}
\end{equation}
where $E(\cdot)$ represents the expectation and $\gamma \in [0,1]$ denotes the discount factor. The temporal difference (TD) target~\cite{lillicrap2015continuous} is as supervision for the value:
\vspace{-1mm}
\begin{equation}
\label{eq:td target}
    q_t=\begin{cases}
    r_{t+1} + \gamma \max_{a}{Q\left(s_{t+1}, a\right)},& \text{if } t < M-1\\
    r_{T},              & \text{otherwise}
    \end{cases}
\vspace{-1mm}
\end{equation}

To minimize the rendering error in step 1, The reward \(r\) at the current step is defined as the negative loss from the coarse DNR loss \(-L_{DNR_c}\), which is further detailed in Sec.~\ref{sec:training}.
And we employ the mean square error (MSE) $L_{MSE}$ to calculate the reinforcement learning loss $L_{RL}$:
\vspace{-2mm}
\begin{equation}
L_{RL}=\sum_{t=1}^{T-1}L_{MSE}(Q(s_{t},a_{t+1}),q_{t})
\vspace{-2mm}
\end{equation}
where the next action $a_{t+1}$ representing the next view is selected using the $\epsilon$-greedy algorithm.

In Step 1, we jointly optimize the DNR model with \(L_{DNR_c}\) and the proposed reinforcement learning-based view selector with \(L_{RL}\). 
Conditioned on the historical and current training states, the RL-based view selector is designed to choose the next best view, guided by the DNR model. 
After rapidly converging on the rasterized images, the RL-based view selector outputs the best \(M\) camera views, reducing the need for the target object with numerous viewpoints.
\vspace{-1.5mm}
\subsection{3D Texture Aggregator}
Vanilla deferred neural rendering employs hierarchical sampling on the UV map to extract neural textures. However, when the number of views is insufficient for optimization, DNR has difficulty capturing the underlying geometry of the object. To address this, we propose a 3D texture aggregator that utilizes depth maps, normal maps, and UV maps to enhance the representational capability for complex topologies.
The 3D texture aggregator revolves around the integration of depth maps, normal maps, and UV maps, all of which play an indispensable role in the intricate representation of 3D objects. 
Depth maps $\mathbf{D}_t$ provide the distance of surfaces from the camera, 
normal maps $\mathbf{N}_t$ encode surface details through normal vectors, 
and UV maps $\mathbf{U}_t$ denote the mapping between 3D coordinates and 2D texture space. 

In the current training iteration $t$, initially we apply the hierarchical sampling strategy $\pi(\cdot)$ on the three maps to extract neural textures respectively: $\{\mathbf{Tex}_D,\mathbf{Tex}_N,\mathbf{Tex}_U\}=\{\pi(\mathbf{D}_t), \pi(\mathbf{N}_t), \pi(\mathbf{U}_t)\}$. These neural textures are then concatenated and processed through a convolutional neural network to form the spatial neural texture $T$. The spatial neural texture then serves as the state of DNR in the view selector and is subsequently sent to the U-Net-like renderer \(R\) to generate high-fidelity images.


\vspace{-1.5mm}

\subsection{Training DNRSelect}
\label{sec:training}

As illustrated in Fig.~\ref{fig:pipeline}, DNRSelect involves two steps. Step 1 jointly trains the proposed view selector and DNR model on the rasterized images. Loss $L_{RL}$ for the RL-based view selector has been exhibited in Sec.~\ref{sec:rl-based}. The DNR model learns an approximate geometric representation with the photometric loss between rendering rasterized images $\hat{I_c}$ and ground truths $I_c$: $L_{DNR_c}(\hat{I_c}, I_c)=L_{MSE}(\hat{I_c}, I_c)$. The total loss in Step 1 can be formulated as:
\vspace{-2mm}
\begin{equation}
L_{step_1}=\lambda_{RL}L_{RL}+\lambda_{DNR_c}L_{DNR_c}
\vspace{-2mm}
\end{equation}
where $\lambda_{RL}$ and $\lambda_{DNR_c}$ denote loss weights respectively.

Step 2 optimizes the coarse DNR model leveraging the ray-traced images of the selected views from Step 1. The vanilla DNR tends to overfit and produce artifacts when trained with a limited set of images. Therefore, we adopt a combination of six different losses for DNR fine-tuning in Step 2, including photometric loss \( L_{DNR_f} \), SSIM loss $L_{SSIM}$, frequency reconstruction loss $L_{FR}$, total variation loss $L_{TV}$, perceptual loss $L_{p}$ and texture regularization loss $L_{reg}$. 
The ground truth ray-traced image is denoted as $I$ and the prediction is $\hat{I}$. 
To better align with human visual perception, we implement $L_{SSIM}$ and $L_{p}$:
\vspace{-1mm}
\begin{equation}
L_{SSIM}(\hat{I},I)=1-SSIM(\hat{I},I)
\vspace{-0.5mm}
\end{equation}
\begin{equation}
L_{p}(\hat{I},I)=\sum_{j}w_{j}||\phi_{j}(\hat{I})-\phi_{j}(I)||
\vspace{-1.5mm}
\end{equation}
where $SSIM$ denotes the structural similarity index measure. In $L_{p}$, we adopt the pre-trained VGG networks to extract features. $\phi_{j}$ is the feature from the $j$-th layer of the pre-trained feature extractor and $w_{j}$ is the weight associated with the $j$-th feature loss term.

To enhance high-frequency details lying in the images, we apply the Fast Fourier Transform $F(\cdot)$ on the images and compute the $L1$ distance as $L_{FR}$:
\vspace{-1.5mm}
\begin{equation}
L_{FR}(\hat{I},I)=||F(\hat{I})-F(I)||
\vspace{-1.5mm}
\end{equation}

We also implement \(L_{TV}(\hat{I})\) to enhance color smoothness and \(L_{reg}(T)\) to regularize the neural textures to small values, ensuring stable optimization.

Eventually, we take the weighted average of the six losses to compute the total loss $L_{step2}$ in Step 2:
\vspace{-1.5mm}
\begin{equation}
\begin{aligned}
L_{step2} = & \lambda_{DNR_f}L_{DNR_f}+\lambda_{SSIM}L_{SSIM}+ \lambda_{p}L_{p}\\
            & 
            +\lambda_{FR}L_{FR}+\lambda_{TV}L_{TV}+\lambda_{reg}L_{reg} 
\end{aligned}
\vspace{-1mm}
\end{equation}
where $\lambda_{DNR_f}$, $\lambda_{SSIM}$, $\lambda_{p}$, $\lambda_{FR}$, $\lambda_{TV}$ and $\lambda_{reg}$ stand for corresponding loss weights respectively.

Under the joint supervision of multiple loss functions, DNRSelect is trained until convergence. 

\section{Experiments}

\subsection{Experimental Settings}

\begin{figure}
    \centering
    \includegraphics[width=1.0\linewidth]{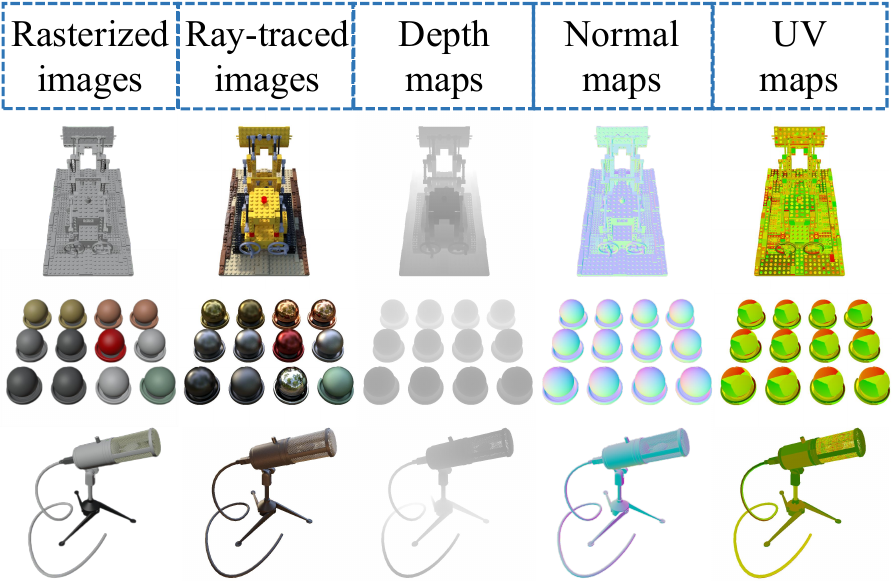}
    \caption{A selection of our preprocessed NeRF-Synthetics dataset is illustrated in the figure.}
    \label{fig:dataset_fig}
    \vspace{-7mm}
\end{figure}

\noindent{\textbf{Dataset and Preprocess.}} 
We evaluate our method on the commonly used dataset NeRF-Synthetics~\cite{mildenhall2021nerf}. 
The dataset consists of 8 synthetic objects, each with 400 images captured from various viewpoints. 
All images are ray-traced and rendered using Blender software at a resolution of 800\(\times\)800.
Images from each object are split into a training set (100 images), a validation set (100 images), and a test set (200 images). 
To generate the UV maps for DNR, we import the 3D objects into Blender, which produces UV maps, rasterized images, depth maps, and normal maps for each camera pose. This CPU-based process is much faster than GPU-intensive ray-traced rendering.
The final synthesized dataset with different modalities is partly shown in Fig.~\ref{fig:dataset_fig}.

\noindent{\textbf{Implementation Details. }}
For each object scene, DNRSelect is trained in two steps to achieve a 3D representation and synthesize new perspectives. 
Step 1 requires 50 epochs, while Step 2 takes 100 epochs. 
Additionally, to prevent overfitting by the U-Net renderer, we employ a data augmentation strategy that includes random flipping and rotation during preprocessing, which is further evaluated in the ablation study. 
The learning rate is initially set to 0.001 and then reduced to 0.0001, using the Adam optimizer~\cite{zhang2018improved} with $(\beta_1, \beta_2) = (0.9, 0.999)$ to update the model parameters. 
For the loss weights in Step 1, \(\lambda_{RL}\) is set to 0.1 and \(\lambda_{DNR_c}\) to 1.0. 
In Step 2, the loss weights are set as follows: \(\lambda_{DNR_f} = 1.0\), \(\lambda_{SSIM} = 0.1\), \(\lambda_{p} = 0.1\), \(\lambda_{FR} = 0.01\), \(\lambda_{TV} = 0.001\), and \(\lambda_{reg} = 0.1\). 
All experiments are conducted on NVIDIA V100 GPUs with PyTorch version 1.10.0.

\vspace{-1.5mm}
\subsection{Main Results}

\begin{figure*}
    \centering
    \includegraphics[width=1.0\linewidth]{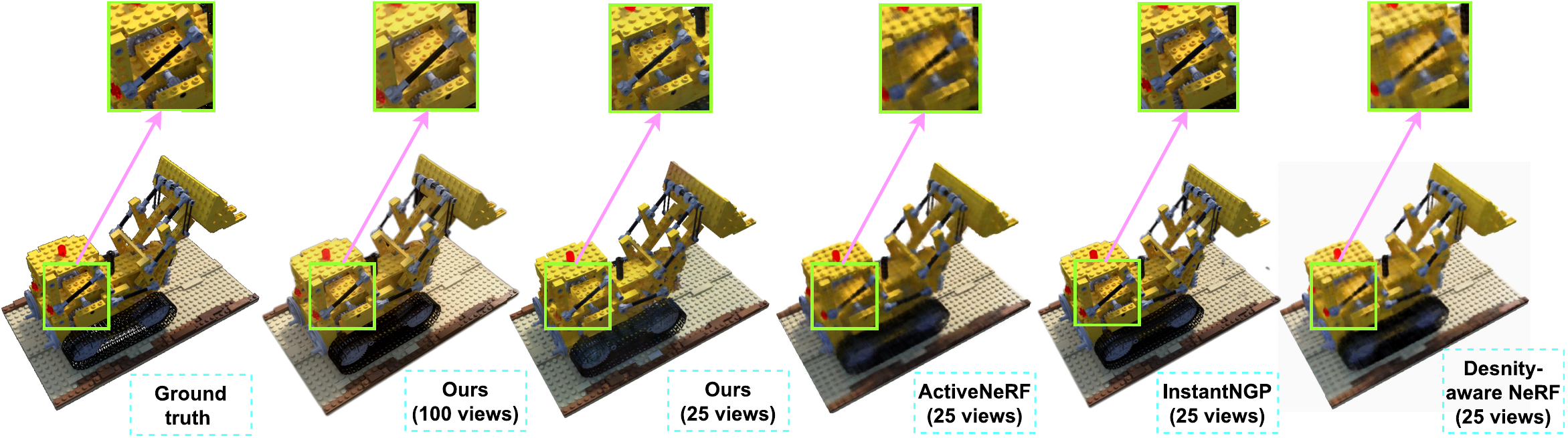}
    \caption{Visualization of the rendering details for our method at 25 and 100 view numbers, compared to other methods at 25 view numbers, alongside the ground truth.
    }
    \label{fig:visual present}
    \vspace{-6mm}
\end{figure*}
\begin{figure}
    \begin{subfigure}{1\linewidth}
        \centering
        \includegraphics[width=\linewidth]{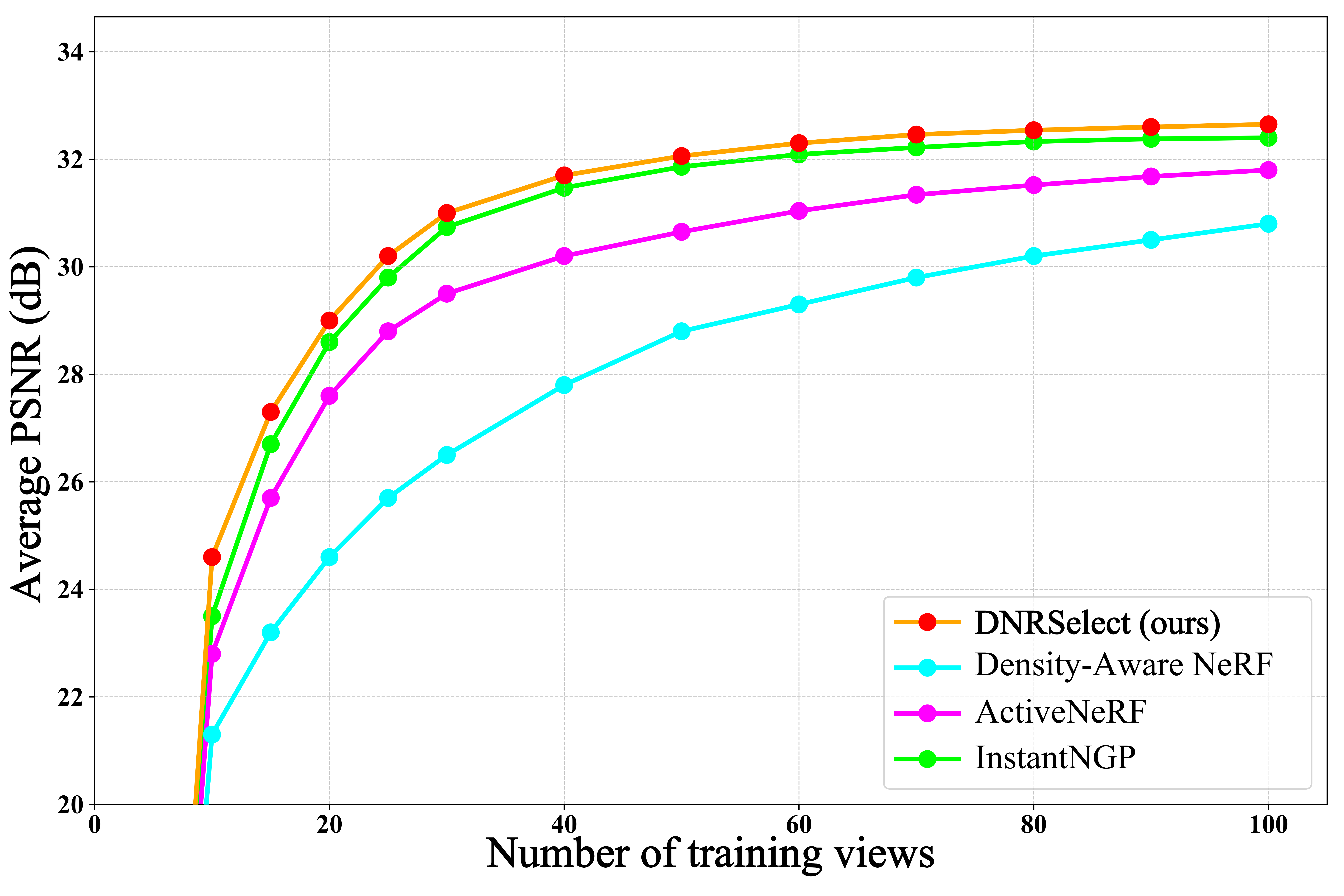}
        \vspace{-6mm}
        \caption{Average PSNR versus the number of training views.}
        \label{fig:6a}
    \end{subfigure}
    \begin{subfigure}{1\linewidth}
        \centering
        \includegraphics[width=\linewidth]{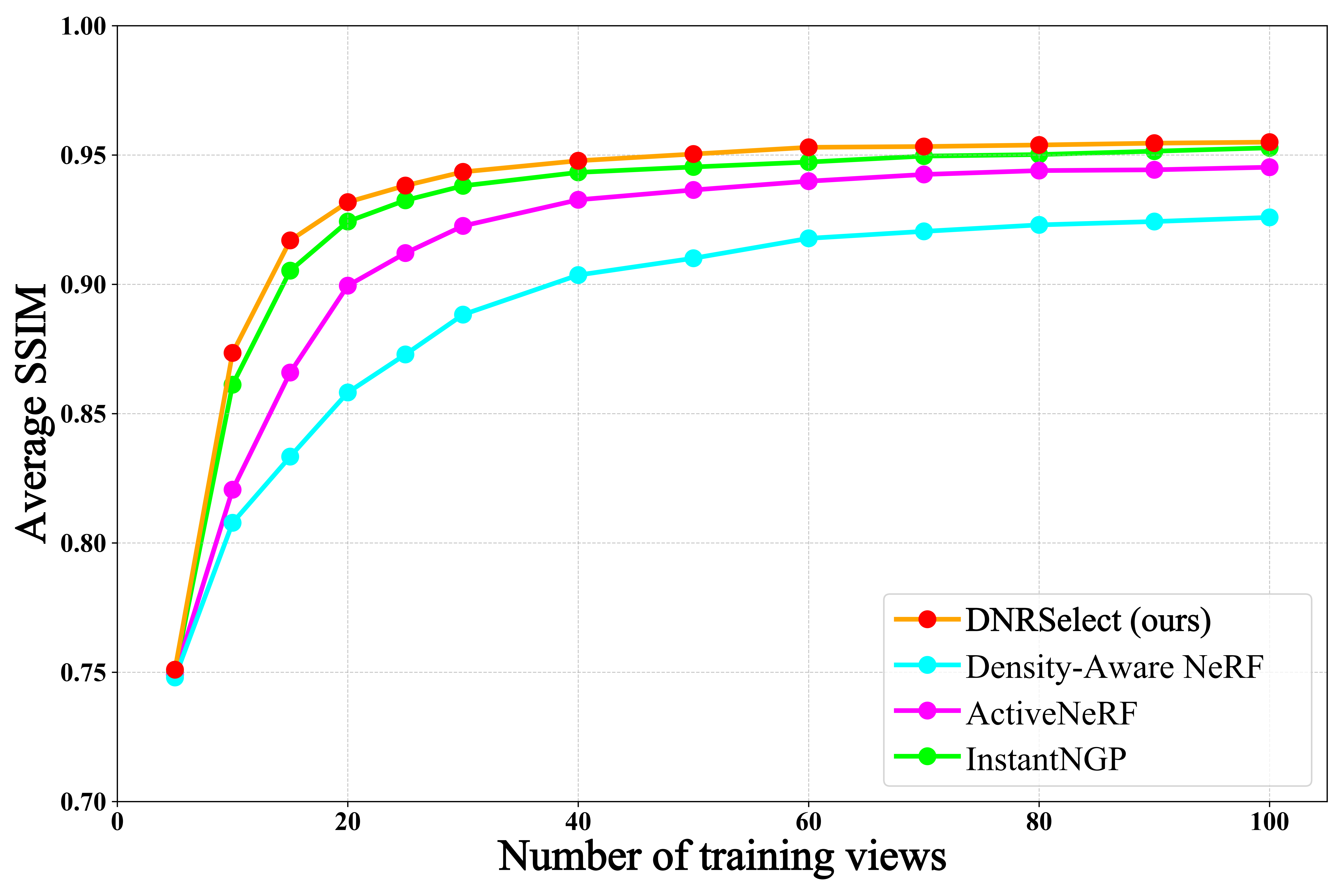}
        \vspace{-6mm}
        \caption{Average SSIM versus the number of training views.}
        \label{fig:6b}
    \end{subfigure}
    \caption{Quantitative comparisons of rendering quality as the number of training views increases, sampled using different view selection methods. Results are evaluated on the NeRF Synthetic dataset.}
    \label{fig:main}
    \vspace{-7mm}
\end{figure}
\begin{figure}
    \centering
    \includegraphics[width=1\linewidth]{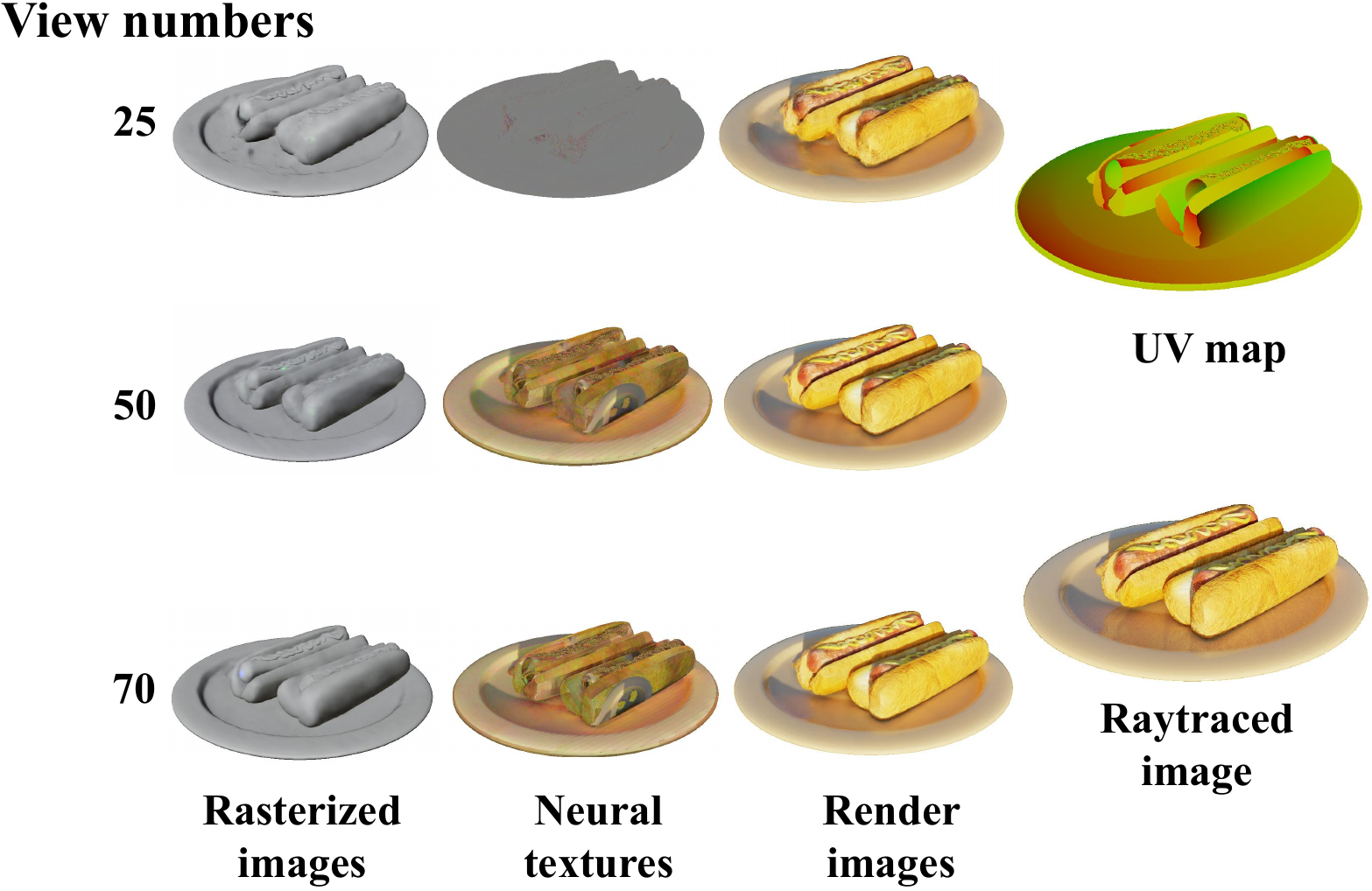}
    \vspace{-2mm}
    \caption{The three columns on the left show results from training at different view numbers. The rasterized image represents the network's predicted output in Step 1, while the ray-traced image is the output from Step 2. The input UV map and the corresponding true ray-traced image are on the right for the current test view numbers and scene.}
    \label{fig:2}
    \vspace{-7mm}
\end{figure}

We conduct a comprehensive comparison of our DNRSelect against several state-of-the-art techniques: ActiveNeRF~\cite{pan2022activenerf}, Density-aware NeRF Ensembles~\cite{10161012}, two uncertainty-based information gain sampling variants, and InstantNGP~\cite{mueller2022instant}. Our process starts with 5 selective views, gradually adding 5 views up to 30, and then 10 views at a time until reaching 100. Quantitative results are shown in Fig.~\ref{fig:main}. We also exhibit the visualization in Fig.~\ref{fig:visual present} and Fig.~\ref{fig:2}. 

Obviously, our results demonstrate that the proposed reinforcement learning-based approach is more compatible with the DNR architecture than other methods, effectively reducing the number of required views while maintaining high rendering quality.

DNRSelect significantly outperforms other rendering methods and nearly reaches the theoretical upper limit. Remarkably, with fewer selected views, it can sometimes exceed this limit, highlighting its efficiency in leveraging optimal views for learning. This is due to the representativeness and diversity of the chosen views, which allow the network to extract more effective features. Our approach handles prediction uncertainties better and ensures more consistent reconstructions across views.

Visual results further support our findings, showing clear improvements in rendering quality as the number of selectable viewing angles increases, consistent with the quantitative data. 
When the number of angles reaches one-quarter of the total, the rendering quality closely matches the ground truth and surpasses that of other methods in detail. Although some artifacts remain, our method captures fine details, achieving high-quality rendering with fewer viewpoints by strategically selecting views that effectively cover the object.

\subsection{Ablation Study}

\begin{table*}[htb]
    \centering
    \caption{In this ablation experiment, we utilize the same hyperparameters and training iterations as described earlier, with training conducted from 25 views. After training, we test the models on the same dataset and record three metrics.}
    \setlength{\tabcolsep}{2pt} 
    \renewcommand{\arraystretch}{1.2} 
    \begin{tabular}{p{3cm} p{3cm} p{3cm} p{3cm} p{3cm} p{0cm}}
    \hline
     \centering\multirow{2}{*}{\textbf{Data augmentation} } & \multicolumn{1}{c}{\textbf{Multiple loss function }}  &  \centering\multirow{2}{*}{\textbf{RL-based view selector}} &  \centering\multirow{2}{*}{\textbf{3D texture aggregator}} & \multicolumn{1}{c}{\textbf{Image test}} \\
    \cline{5-5}
    &  \centering\textbf{supervision}& & &\textbf{ PSNR} / \textbf{SSIM} / \textbf{LPIPS} \\
    \hline
         \centering \textcolor{red}{\usym{2715}} & \centering $\checkmark$ & \centering $\checkmark$ & \centering $\checkmark$ & \centering 26.64 / 0.8179 / 0.0542 & \\
         \centering $\checkmark$ & \centering \textcolor{red}{\usym{2715}} & \centering $\checkmark$ & \centering $\checkmark$ & \centering 27.19 / 0.8296 / 0.0497 & \\
         \centering $\checkmark$ & \centering $\checkmark$ & \centering \textcolor{red}{\usym{2715}} & \centering $\checkmark$ & \centering 27.14 / 0.8314 / 0.0498 & \\
         \centering $\checkmark$ & \centering $\checkmark$ & \centering $\checkmark$ & \centering \textcolor{red}{\usym{2715}} & \centering 27.84 / 0.8470 / 0.0503& \\
         \centering $\checkmark$ & \centering $\checkmark$ & \centering $\checkmark$ & \centering $\checkmark$ & \centering \textbf{28.12} / \textbf{0.8506} / \textbf{0.0479} & \\
    \hline
    \end{tabular}
    \label{tab:my_label}
    \vspace{-4mm}
\end{table*}

DNRSelect represents a significant advancement over the classic DNR network, incorporating a range of innovative enhancements such as data augmentation, and multi-loss function supervision. 
More importantly, We introduce a novel reinforcement learning-based view selector and a 3D texture aggregator. 
To rigorously validate the effectiveness of these improvements, we conduct an ablation study by systematically removing each mechanism and assessing its impact on rendering quality under controlled conditions. 

As shown in Tab.~\ref{tab:my_label}, the results demonstrate that all three proposed mechanisms improve rendering performance, each to varying degrees. Data augmentation enhances the DNR framework's generalization, allowing it to better adapt to diverse inputs. Joint supervision with multiple loss functions optimizes texture detail beyond simple RGB absolute loss. The RL-based view selector reduces noise by choosing the most informative views, and the proposed aggregator strategy enhances the network's ability to model 3D objects accurately. Together, these innovations enable DNRSelect to achieve superior rendering quality, highlighting the originality and practicality of our approach.
\vspace{-2mm}
\subsection{Comparison on Aggregator Strategy}
\vspace{-1mm}
We conduct two experiments to examine the impact of different aggregator strategies and data combinations on neural rendering performance. The first experiment compares the vanilla strategy, which merges depth, normal, and UV maps into a single 9-channel input matrix for early integration, with our proposed strategy, which separately processes neural textures from each data type (\(\mathbf{Tex}_D\), \(\mathbf{Tex}_N\), \(\mathbf{Tex}_U\)) to enhance feature extraction. The second experiment evaluates various data combinations: using the UV map alone, UV map with depth map, UV map with normal map, and the combination of all three. All experiments use consistent hyperparameters and training iterations.

\begin{figure}
    \begin{subfigure}{1\linewidth}
        \centering
        \includegraphics[width=\linewidth]{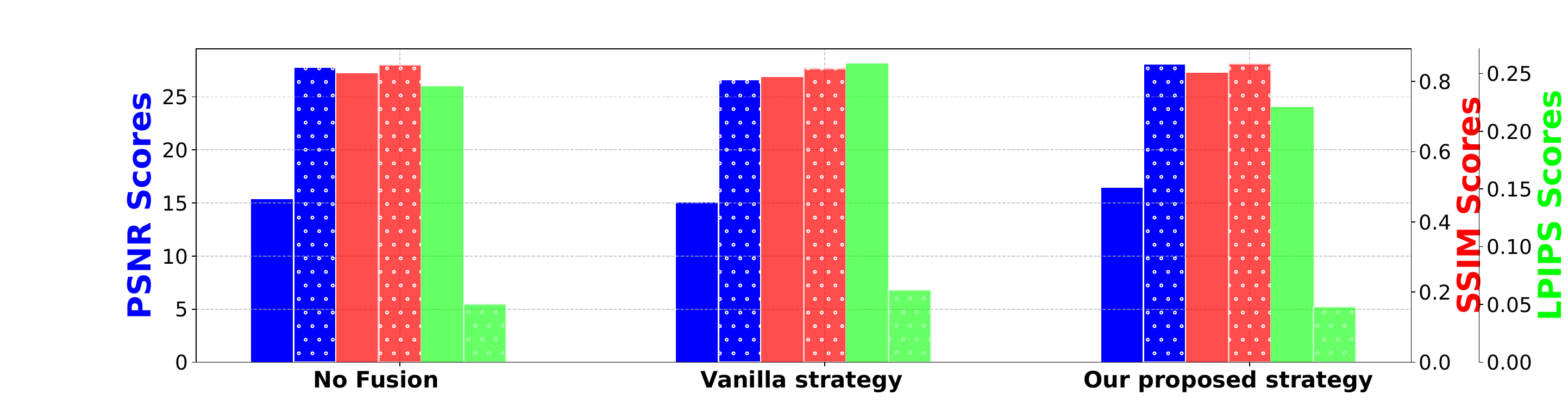}
        \vspace{-6mm}
        \caption{Comparison on data aggregator strategy.}
        \label{fig:7a}
    \end{subfigure}
    \begin{subfigure}{1\linewidth}
        \centering
        \includegraphics[width=\linewidth]{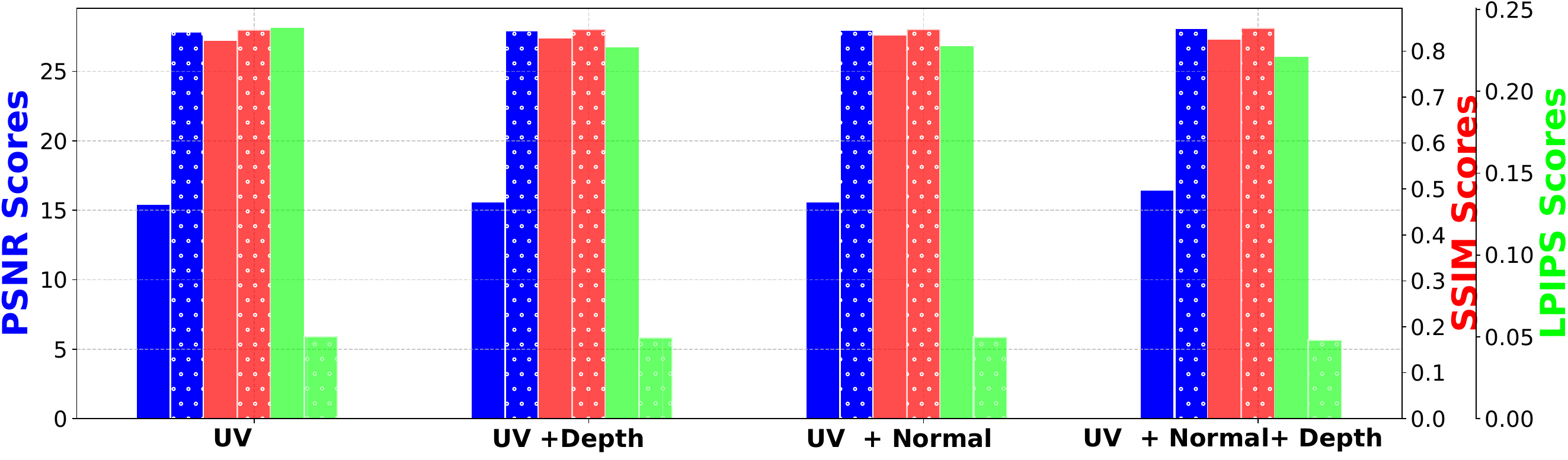}
        \vspace{-6mm}
        \caption{Comparison on different data modalities aggregator.}
        \label{fig:7b}
    \end{subfigure}
    \begin{subfigure}{1\linewidth}
        \centering
        \includegraphics[width=\linewidth]{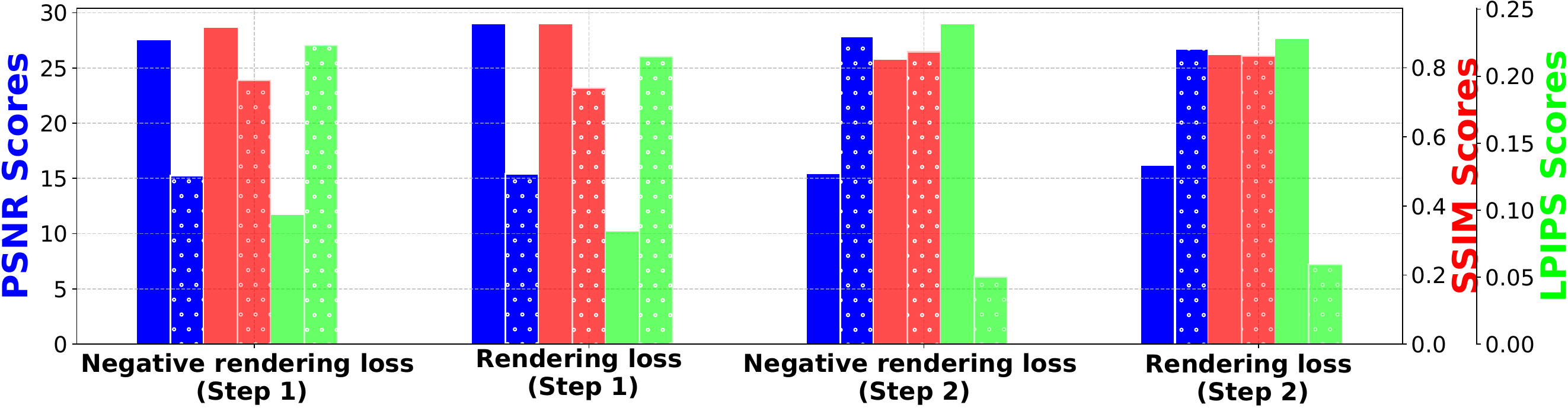}
        \includegraphics[width=\linewidth]{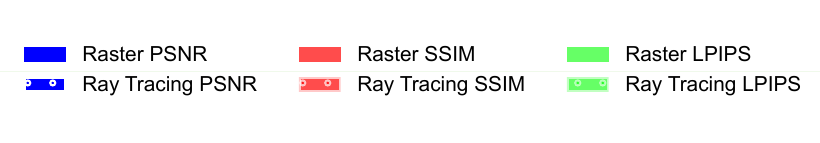}
        \vspace{-6mm}
        \caption{Comparison on reward strategy.}
        \label{fig:7c}
    \end{subfigure}
    \label{fig:com}
    \vspace{-4mm}
    \caption{Quantitative comparisons of rendering quality across various strategies, evaluated using PSNR, SSIM, and LPIPS. }
    \vspace{-7mm}
\end{figure}

As illustrated in Fig.~\ref{fig:7a}, The results show that the proposed aggregator strategy significantly outperforms the original approach, which often performs worse than scenarios without any data fusion. This suggests the original method struggles to capture inter-modal correlations, hindering the network's learning. In contrast, the proposed strategy processes each modality separately before combining their neural textures as high-dimensional features, reducing data inconsistencies and enabling the network to extract more meaningful information, resulting in superior performance.

Further analysis in Fig.~\ref{fig:7b} reveals that combining all three modalities (depth maps, normal maps, and UV maps) achieves the best performance, as the additional data helps the network better understand the 3D object's geometric structure, resulting in more refined textures. However, using depth and normal maps increases data collection and processing time. Notably, the network still performs well without any fusion strategy, indicating that the choice of aggregator should balance optimal rendering quality with processing efficiency based on the specific context.

\vspace{-2mm}
\subsection{Comparison on Reward Strategy}
\vspace{-1mm}
We investigate the impact of different reinforcement learning reward strategies on the view selection process in the DNRSelect network, understanding that effective reinforcement learning relies heavily on its reward mechanisms. 
Our analysis shows that using the negative value of the rendering loss as a reward encourages the selection of simpler views, boosting overall performance. 

The results in Fig.~\ref{fig:7c} show that using the negative rendering loss as a reward is more effective for DNR. This strategy aligns the objectives of the view selector with the rendering network and helps prioritize views that enhance the network's learning. It is especially beneficial in the early stages of rendering, where capturing rough geometric features is more important than detailed rasterization.
\vspace{-1mm}
\section{conclusion}
\vspace{-1mm}
To reduce the reliance on extensive ray-traced images, we propose DNRSelect for active deferred neural rendering. By integrating a reinforcement learning-based view selector, DNRSelect actively identifies optimal views using readily available rasterized images, thereby minimizing the need for computationally expensive ray-traced images. Additionally, we introduce a 3D texture aggregator that enhances spatial awareness and geometric consistency. 
We hope this research paves the way for more adaptive approaches in future deferred neural rendering exploration.

\bibliography{IEEEabrv}
\bibliographystyle{ieeetr}

\end{document}